\documentclass[sigconf, nonacm]{acmart}

\setcopyright{none}
\renewcommand\footnotetextcopyrightpermission[1]{}
\AtBeginDocument{%
  }
\usepackage{booktabs}
\usepackage[table]{xcolor}
\definecolor{proprow}{gray}{0.88}
\begin{document}

\title{Multi-Label Proportion Learning for Sea-Ice Type Prediction}

\author{Samira Alkaee Taleghan}
\affiliation{%
  \institution{University of Colorado Denver}
  \city{Denver}
  \state{Colorado}
  \country{USA}}
\email{samira.alkaeetaleghan@ucdenver.edu}

\author{Younghyun Koo}
\affiliation{%
  \institution{National Snow and Ice Data Center (NSIDC), CIRES, University of Colorado Boulder}
  \city{Boulder}
  \state{Colorado}
  \country{USA}}
\email{younghyun.koo@colorado.edu}

\author{Andrew P. Barrett}
\affiliation{%
  \institution{National Snow and Ice Data Center (NSIDC), CIRES, University of Colorado Boulder}
  \city{Boulder}
  \state{Colorado}
  \country{USA}}
\email{andrew.barrett@colorado.edu}

\author{Farnoush Banaei-Kashani}
\affiliation{%
  \institution{University of Colorado Denver}
  \city{Denver}
  \state{Colorado}
  \country{USA}}
\email{farnoush.banaei-kashani@ucdenver.edu}

\renewcommand{\shortauthors}{Alkaee Taleghan et al.}


\begin{abstract}

Sea-ice type prediction is important for climate monitoring, maritime navigation, and decision-making in polar regions. The main source of label data for this task is the ice chart, produced manually by ice analysts who interpret satellite imagery to delineate ice zones into polygons. Each polygon is assigned total sea-ice concentration and partial concentrations of the ice types present within the polygon. Although ice charts are valuable, their production is labor-intensive and expensive, motivating recent efforts to automate the process using deep learning. However, deep learning models require patch-level (or pixel-level) label data for training, while ice charts provide only polygon-level annotations. As a workaround, supervised approaches often create approximate patch-level labels from polygon-level ice chart labels by assigning each sample the dominant ice type of its parent polygon. This approach enables supervised training but creates an ill-posed learning problem with intrinsically approximate solution. In this paper, we redefine sea-ice type prediction as a weakly supervised multi-label proportion learning problem to be able to directly use the polygon-level ice chart labels and avoid unnecessary label approximation for improved prediction accuracy. To address this problem, we propose a two-module framework where first Multiple Instance Learning (MIL) is used for water--ice classification, and then a multi-label proportion learning (MLPL) is introduced for ice-type composition prediction. We further extend this framework with a multimodal model that integrates SAR imagery with AMSR2 brightness temperatures and ERA5 reanalysis data through modality-guided auxiliary regularization. Evaluated on the AI4Arctic dataset, the SAR-only model reduces MAE by 14.5\% and more than doubles mean ice-class F1 over the best supervised baseline. The multimodal model further reduces MAE by 21.5\% and raises mean F1 by 41.2\% over the SAR-only model, and by 52.7\% over the supervised multimodal baseline.

\end{abstract}




\maketitle

\section{Introduction}
Sea-ice type prediction is important for climate monitoring, numerical modeling, maritime navigation, and decision-making in polar regions. Because sea-ice type reflects ice thickness, age, strength, and navigability, reliable type information is essential for understanding polar climate processes and supporting safe activity in ice-covered waters \cite{Vihma, taleghan2024semisupervised}. The main source of sea-ice type information is the ice chart. Ice charts are produced manually by expert ice analysts, who interpret multi-sensor satellite imagery and auxiliary data sources using domain expertise and visual inspection to delineate ice zones into polygons. Synthetic Aperture Radar (SAR) is especially important in this process because it provides high-resolution observations and can operate through clouds and polar darkness \cite{ Dierking2013}. Despite their value, ice charts are labor-intensive and expensive to produce, motivating efforts to automate sea-ice type prediction using deep learning.

\begin{figure}[!t]
    \centering
    \includegraphics[width=\columnwidth, height=2in]{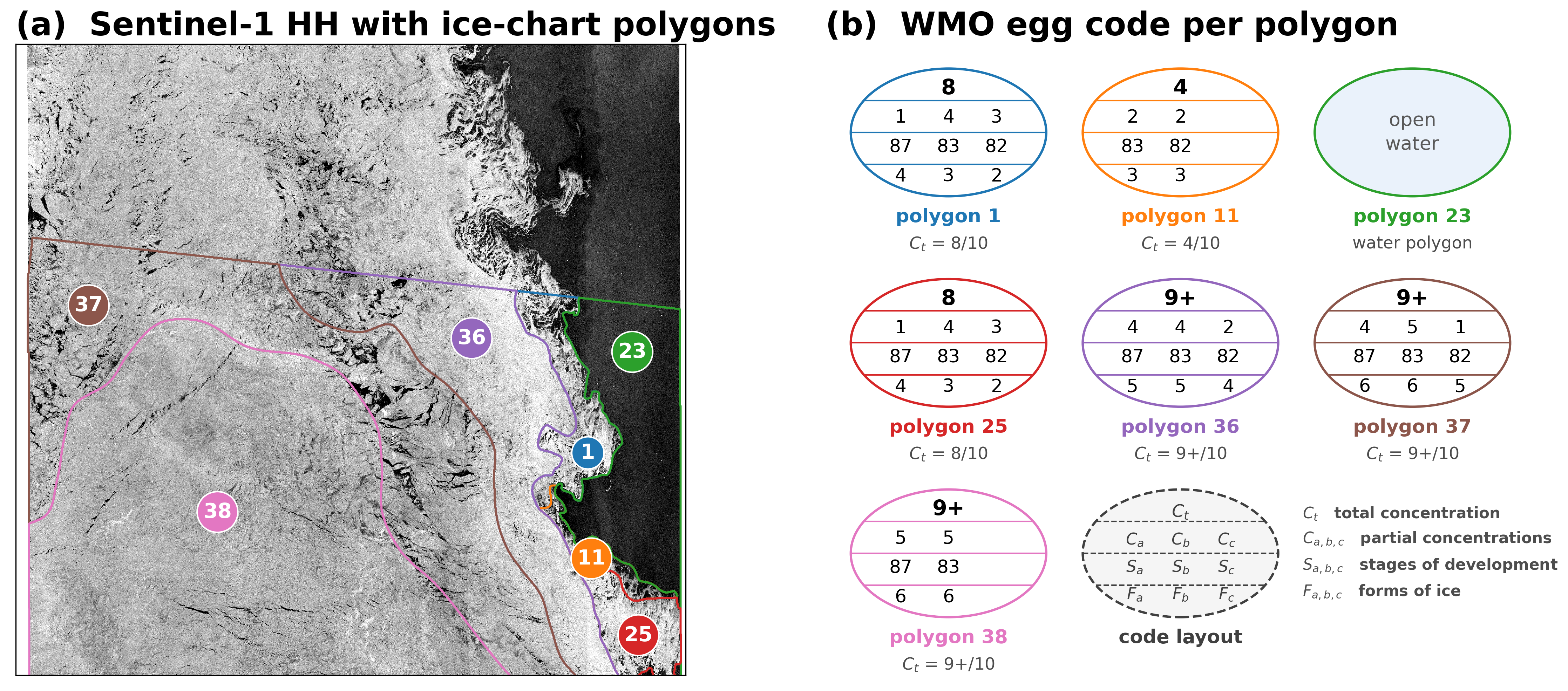}
    \caption{Ice-chart polygons and their WMO egg codes. (a) Sentinel-1A EW HH
scene, 14 October 2021 07:12\,UTC, Fram Strait north-west of Svalbard, DMI ice chart.
(b) SIGRID-3 egg code per polygon.}
\Description{A grayscale SAR satellite image with seven numbered colored polygon outlines drawn over sea-ice regions, alongside seven oval egg-code diagrams. Each oval lists the polygon's total ice concentration on top, with partial concentrations, stages of development, and floe-size codes in rows below.}
    \label{fig:icechart}
\end{figure}

In operational ice charting, analysts manually draw polygons over regions of relatively homogeneous ice conditions and assign each polygon a set of standardized codes following the World Meteorological Organization (WMO) "egg code" and the SIGRID3 data format \cite{NSIDC}. Rather than assigning a single ice type per polygon, the egg code records the total sea ice concentration (SIC) (the fraction of ice cover from 0–100\%) along with the partial concentrations for up to three co-occurring ice types, each characterized by its stage of development (SOD), a proxy for ice thickness, and floe size (FLOE). Figure~\ref{fig:icechart} shows an example of ice-chart polygons produced by the Danish Meteorological Institute (DMI) overlaid on a Sentinel-1 SAR scene, together with the WMO egg code assigned to each polygon. Importantly, these labels describe the aggregate composition of the entire polygon, not the class identity of each individual pixel or patch. Conditions within a polygon may vary substantially, and mixed polygons can contain several ice types with different partial concentrations.

Recent advances in deep learning have opened new opportunities to automate sea-ice type prediction and support ice-chart generation\cite{Li2024, jalayer2025, Taleghan2025icefmbench}. By extracting spatial and textural features from SAR imagery, deep learning models can overcome the constraints of manual interpretation, enabling more scalable and timely ice charting. However, training such models requires labeled data at the pixel or patch level — labels that are rarely available for sea ice; ice charts provide only polygon-level annotations. As a result, supervised approaches convert polygon annotations into pixel- or patch-level labels by assigning each sample the dominant ice type of its parent polygon, thereby reducing the egg code’s multi-label proportional structure to a single class \cite{park2020,boulze2020,kruk2020}. This workaround enables supervised training, but it creates an ill-posed learning problem: dominant-label reduction discards partial concentration information, introduces label noise in mixed polygons, and ignores the uncertainty inherent in training from heterogeneous polygon-level annotations.

The mismatch between the polygon-level proportion annotations available in ice charts and the pixel-level labels required by standard supervised methods defines the core challenge of this work. Rather than inferring noisy dominant-class labels, we preserve the original structure of ice-chart supervision by treating each polygon as a bag of unlabeled image patches whose collective composition is described by label proportions. This setting naturally aligns with Multiple Instance Learning (MIL), where supervision is provided for bags rather than individual instances \cite{Dietterich1997, Maron1998}, and with Learning with Label Proportions (LLP), where aggregated instance predictions are trained to match known bag-level class proportions \cite{Quadrianto2009}. Because ice-chart polygons may contain multiple co-occurring ice types, the resulting task is both multi-label and proportion-based. This formulation avoids dominant-class assumptions and directly leverages the full annotation richness of operational ice charts. 
We propose a two-module framework that mirrors the structure of the ice-charting task. The first module performs coarse water--ice classification using a MIL formulation. The second module estimates the ice-type composition of ice polygons using multi-label proportion learning (MLPL). In the second module, the core SAR-only model uses evidential Dirichlet aggregation, allowing patches to contribute variable evidence to polygon-level proportion estimates rather than forcing every patch to contribute equal softmax mass. This is important because boundary patches, ambiguous textures, and mixed regions should not influence the bag-level estimate in the same way as visually informative ice-type patterns. As an extension, we introduce a multimodal auxiliary-regularized architecture that integrates SAR imagery with AMSR2 brightness temperatures and ERA5 atmospheric reanalysis fields through AMSR2 ice-concentration consistency and ERA5 environmental-compatibility regularization.

The main contributions of this work are: (1) we formulate the polygon-level labeling stage of operational sea-ice
charting as a weakly supervised problem that preserves partial concentrations instead of reducing each polygon to a dominant class, and introduce a two-module framework that combines binary MIL with multi-label proportion learning: first identifying ice-covered polygons, then predicting fine-grained multi-label ice-type proportions directly from polygon-level concentration labels; (2)  we adapt evidential Dirichlet aggregation to bag-level sea-ice proportion learning, allowing patches to contribute variable evidence rather than equal softmax mass; and (3) we extend the framework with modality-guided auxiliary regularization,
incorporating AMSR2 ice-concentration consistency and ERA5 environmental
compatibility. Additionally, we investigate bag-size calibration and uncertainty-weighted Dirichlet training as extensions that further improve proportion accuracy, and analyze the trade-off between patch-selection budget and prediction performance.

The remainder of the paper is organized as follows: Section 2 reviews related work, Section 3 presents the methodology, Section 4 reports the experiments and results, and Section 5 concludes the paper.
\section{Related Work} 

\subsection{Sea Ice Type Classification}
Supervised sea ice classification methods aim to automatically identify and distinguish ice types from satellite imagery using labeled training data.
Early CNN-based studies demonstrated the feasibility of deep learning for sea ice classification. Li et al. \cite{Li2017} applied CNNs to Gaofen-3 SAR imagery to distinguish ice from open water, and Boulze et al. \cite{boulze2020} used CNNs on Sentinel-1 data to classify ice types with higher accuracy than traditional methods. Deeper architectures such as AlexNet \cite{Krizhevsky}, VGG16 \cite{Simonyan}, ResNet \cite{He}, and DenseNet \cite{Huang} further improved performance through hierarchical feature extraction and residual learning. Xu et al. \cite{Xu2017} fine-tuned AlexNet for SAR-based ice–water classification, while Khaleghian et al. \cite{Khaleghian2021} showed VGG16’s superiority with augmented training. Specialized, attention-based, and multi-sensor architectures further improved sea-ice classification from SAR data \cite{Song2018,Lyu2022,Zhang2021,kruk2020,Han2022,Chen2023}. IceBench~\cite{alkaeetaleghan2025icebench} provided a standardized benchmark for sea-ice type classification using representative CNN models and common metrics.
Despite these advances, supervised methods often reduce mixed polygons to single labels. In contrast, our work learns directly from polygon-level multi-label proportions, preserving mixed ice-type composition.

\subsection{Multi Instance Learning}

Multiple Instance Learning (MIL) is a weakly supervised framework in which labels are assigned to bags of instances rather than individual samples \cite{Dietterich1997,Maron1998}. Under the classical assumption, a positive bag contains at least one positive instance, while a negative bag contains only negative instances. MIL has since been extended to diverse applications in computer vision, remote sensing, and medical imaging \cite{Amores2013,Carbonneau2018,7812612}.
Traditional MIL uses models such as kNN, SVMs, and decision trees \cite{wang2000,andrews2002,chevaleyre2001}, while deep MIL typically extracts instance features with CNNs and aggregates them at the bag level. Mean and max pooling can respectively dilute informative instances or ignore broader context; attention-based MIL addresses this by learning instance importance weights \cite{Ilse2018}.
Attention-based MIL has progressed from global self-attention models such as TransMIL \cite{Shao2021TransMIL} to local-context approaches such as CAMIL \cite{fourkioti2023camil}. Other variants introduce clustering, distribution-guided scoring, pseudo-bag distillation, graph-based spatial modeling, and multi-task or multimodal learning \cite{clam,Li2023DGMIL,Li2021DTFD,tu2019gnnmil,pal2022,li2021,ren2025,luo2025}. In sea-ice mapping, Alter-CNN uses polygon-level proportions to infer pixel-level ice/water labels \cite{li2015altercnn}, whereas our method directly predicts polygon-level ice-type proportions.
Multi-Instance Multi-Label Learning (MIML) extends MIL to bags associated with multiple labels. Zhou et al. \cite{ZHOU20122291} formalized this setting and introduced methods such as MimlBoost and MimlSvm, while later deep approaches such as MIML-FCN+ incorporated fully convolutional networks \cite{yang2017}.

MIL is used in remote sensing to learn from coarse labels by treating images or regions as bags and pixels, spectra, patches, or superpixels as instances. Applications include hyperspectral target characterization \cite{Jiao2018MIHE}, attention-based scene classification \cite{Li2020DeepMILScene}, superpixel-based hyperspectral classification \cite{Huang2024MSMIL}, and multi-resolution Earth-observation learning \cite{Early2023MultiRes}. However, classical MIL typically assumes binary or categorical bag-level labels and does not directly represent the fractional composition of multiple co-occurring classes. Although MIML extends MIL to multiple labels per bag, it generally models label presence rather than their associated proportions.
\subsection{Learning with Label Proportion}

Learning With Label Proportions (LLP) extends weak supervision by providing class proportions for each bag rather than instance labels \cite{Quadrianto2009,Rueping2010}. Early methods adapted conventional classifiers, such as Proportional SVM ($\propto$SVM), to enforce bag-level proportion constraints \cite{yu2013}. More recent approaches include EasyLLP, which reconstructs instance-level losses from bag proportions \cite{busafekete2023}, LLP-BP, which refines pseudo-instance labels using belief propagation \cite{havaldar2024llpbp}, and theoretically grounded formulations under mutual contamination \cite{scott2020}.
Deep and generative LLP methods have also been proposed, including LLP-GAN \cite{llpgan} and the Batch Averager framework, which averages instance predictions within each bag and matches them to known proportions using KL divergence \cite{ardehaly2017}. In sea-ice applications, MIPL-Ice \cite{alkaeetaleghan2022} introduced an initial LLP-based framework for polygon-level multi-class classification, but did not fully develop or evaluate the multi-instance proportion-learning setting considered here.
In remote sensing, LLP has been applied to SAR classification, crop mapping, and satellite imagery using aggregate proportion labels \cite{Ding2017LLP,LaRosa2022LLPCo,LaRosa2023CropLLP,RamosPollan2025LLPEO}. However, LLP is commonly formulated for proportions of mutually exclusive instance classes and does not explicitly address settings where multiple co-occurring labels and their partial concentrations are jointly represented at the bag level.

\section{Methodology}

\begin{figure}[tb]
    \centering
    \includegraphics[width=\linewidth]{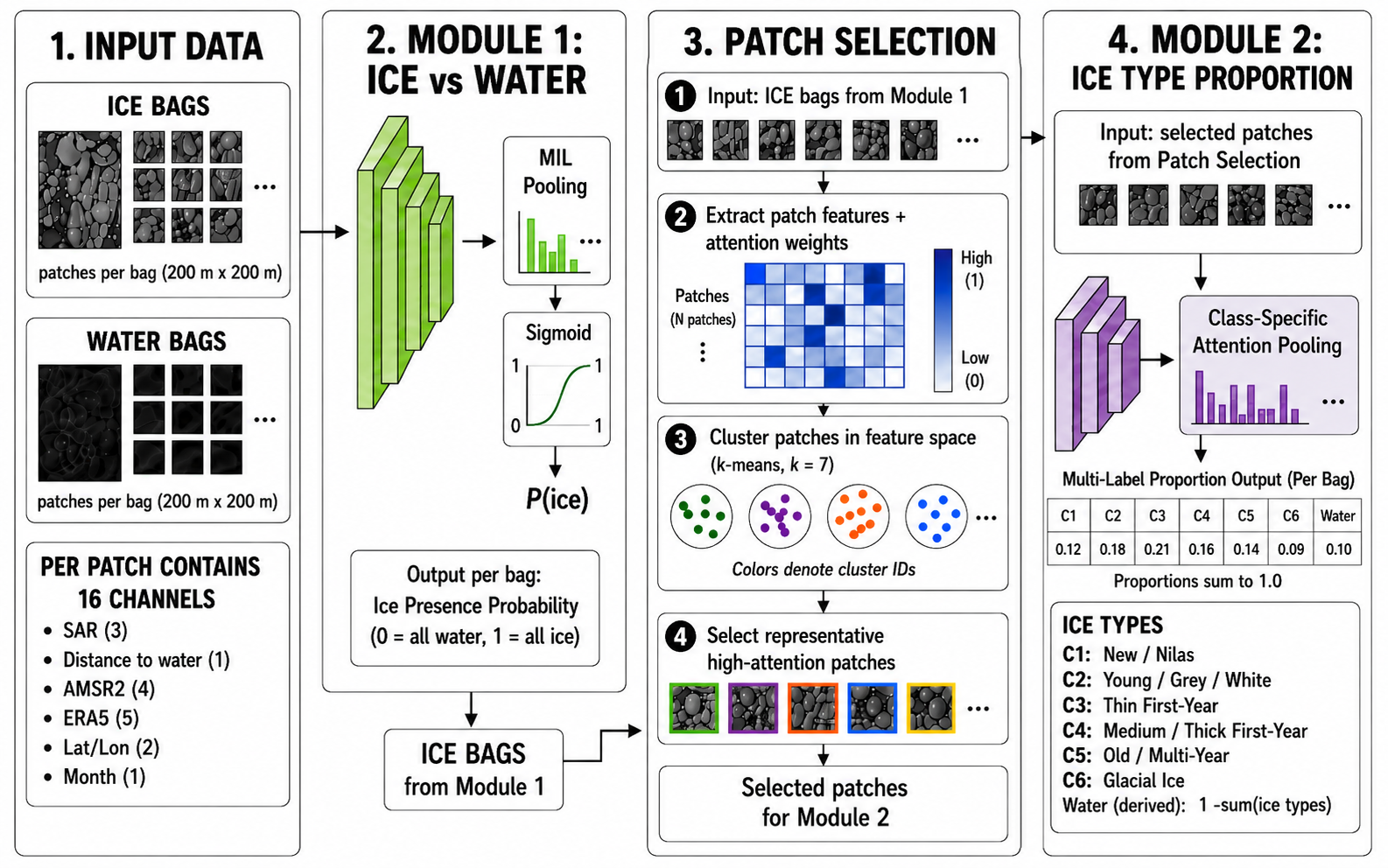}
    \caption{Overview of the proposed model flow structure.}
    \Description{A four-stage pipeline diagram. Stage 1: input data consisting of ice bags and water bags of image patches with 16 channels including SAR, AMSR2, and ERA5. Stage 2: Module 1 classifies ice versus water using MIL pooling and outputs an ice probability per bag. Stage 3: patch selection extracts patch features and attention weights, clusters patches in feature space with k-means, and selects representative high-attention patches. Stage 4: Module 2 takes the selected patches and predicts multi-label ice-type proportions per bag over seven classes using class-specific attention pooling, with proportions summing to one.}
    \label{fig:model_flow}
    \vspace{-3mm}
\end{figure}

Our approach mirrors manual ice charting: first separating open water from ice-covered regions, then estimating ice-type composition within the ice. This motivates a two-module weakly supervised architecture. The first treats water–ice discrimination as a binary MIL problem, while the second uses Multi-Label Proportion Learning (MLPL) to estimate up to three co-occurring ice types from polygon-level partial concentrations. Both modules use only polygon-level supervision, with no patch- or pixel-level labels. Figure~\ref{fig:model_flow} shows the overall model flow.
\subsection{Problem Formulation}

Formalizing this setup, we represent each annotated polygon as a weakly supervised
bag of image patches. For polygon $i$, the bag is defined as
$\mathcal{B}_i = \{x_{i1}, \ldots, x_{iN_i}\}$, where
$x_{ij} \in \mathbb{R}^{H \times W \times M}$ denotes the $j$-th extracted
patch, $M$ is the number of input channels, and $N_i$ is the number of valid
patches in the polygon.
The patches are generated on a regular grid over the polygon bounding box and are
retained only when the fraction of patch pixels falling inside the polygon exceeds a predefined coverage threshold. Because ice-chart
polygons are irregular, some retained patches may intersect polygon boundaries.
For these boundary patches, pixels outside the polygon are imputed from valid pixels
inside the same polygon, so that background or zero-padding artifacts are not
introduced into the MIL bag. 

Each training bag is constructed from patches contained within an ice-chart polygon. The corresponding polygon-level water/ice type provides the weak supervision signal for the ice--water classification module. The objective of this module is to learn a binary MIL classifier that aggregates patch-level evidence to predict a bag-level water--ice label:
\begin{equation}
    \hat{y}_i^{\mathrm{bin}} = f_{\theta}^{\mathrm{ice}}(\mathcal{B}_i),
\end{equation}
where $y_i^{\mathrm{bin}} \in \{0,1\}$ denotes the polygon-level water-ice label, with
$y_i^{\mathrm{bin}}=0$ for open water and $y_i^{\mathrm{bin}}=1$ for sea ice. 

For polygons labeled as sea ice, the egg-code provides a more detailed description of
the ice composition. Specifically, it records up to three dominant ice types
$(S_A, S_B, S_C)$, listed in descending order of thickness, together with their 
corresponding partial concentrations $C_A, C_B, C_C \in \{10,20,\ldots,100\}$
and the total ice concentration $CT$. Therefore, each ice polygon is treated as a
multi-label proportion bag: more than one ice type is present, and the label
specifies the concentration of each present type. From these quantities, we construct a bag-level proportion vector
$\mathbf{y}_i = [y_{i1}, y_{i2}, \ldots, y_{iK}] \in \Delta^{K-1}$
over $K=7$ classes: six ice stages and open water. The six ice classes are
New/Nilas, Young/Grey/White, Thin First-Year, Medium/Thick First-Year,
Old/Multi-Year, and Glacier ice. The open-water component inside an
ice polygon is recovered as the complement of the total assigned ice concentration,
$y_{i,\mathrm{water}} = 1 - \sum_{c=1}^{K-1} y_{ic} = 1 - CT$. 

The second module operates on ice polygons and learns to estimate the fine-grained
multi-label ice-type composition vector $\mathbf{y}_i$ from the bag of patches. Its objective is
to learn a proportion estimator
\begin{equation}
    \hat{\mathbf{y}}_i = f_{\theta}^{\mathrm{prop}}(\mathcal{B}_i),
    \qquad
    \hat{\mathbf{y}}_i \in \Delta^{K-1},
\end{equation}
that matches the polygon-level concentration vector $\mathbf{y}_i$. This corresponds to a multi-label proportion learning (MLPL) setting: the supervision signal is provided only as a polygon-level vector of ice-type partial concentrations, while the class identity of each individual patch remains latent.
Because ice-chart polygons vary substantially in size and shape, the number of patches
per bag $N_i$ is naturally variable. The model must therefore aggregate information
over bags of different cardinalities rather than assuming a fixed number of instances.

\subsection{Ice-Water Classification Module}

The ice-water classification module is a binary MIL classifier that learns to
separate ice-containing bags from open-water bags. For this coarse water-ice
decision, we use the lower-resolution patch set produced by coarsening the raw
40 m SAR grid by a factor of five. This is appropriate because the water-ice boundary is a coarse, visually dominant SAR signal, whereas finer spatial detail is mainly needed for the later ice-type proportion prediction module.

Given the bag defined above, each patch $x_{ij}$ is passed through a convolutional feature
extractor to obtain a patch-level embedding
$\mathbf{h}_{ij}^{\mathrm{ice}}=\phi_{\theta}^{\mathrm{ice}}(x_{ij})$, where
$\phi_{\theta}^{\mathrm{ice}}$ denotes the feature extractor of the ice-water
classification module. The backbone consists of a ResNet feature extractor followed
by a feature projection layer. For standard three-channel SAR input, the ResNet
backbone is used in its conventional form. For multimodal input, the first
convolutional layer is modified to accept the larger number of input channels. The
pretrained weights corresponding to the first three channels are retained, while
the additional channel weights are initialized separately and learned during
training.
For each patch embedding $\mathbf{h}_{ij}$, an instance-level classifier predicts
the probability that the patch contains ice,
$p_{ij}^{\mathrm{ice}}=\sigma(g_{\theta}(\mathbf{h}_{ij}))$, where $g_{\theta}$ is a multilayer classifier and $\sigma(\cdot)$ is the sigmoid
activation function.

We use attention-guided top-$k$ MIL aggregation. 
A learnable two-layer attention network with a Tanh nonlinearity assigns an 
importance score to each patch:
\begin{equation}
    a_{ij}
    =
    \frac{
        \exp\!\left(\mathbf{w}_2^\top \tanh(\mathbf{W}_1 \mathbf{h}_{ij})\,/\,\tau\right)
    }{
        \sum_{m=1}^{N_i}
        \exp\!\left(\mathbf{w}_2^\top \tanh(\mathbf{W}_1 \mathbf{h}_{im})\,/\,\tau\right)
    },
\end{equation}
where $\mathbf{W}_1 \in \mathbb{R}^{128 \times D}$ and $\mathbf{w}_2 \in \mathbb{R}^{128}$ 
are the learnable parameters of the attention network, and $\tau$ is the attention 
temperature. The aggregation then selects the $k$ patches with the largest attention 
weights, \(\mathcal{T}_i =
\mathrm{TopK}\!\left(\{a_{ij}\}_{j=1}^{N_i}, k\right)\), where \(k=10\)
in our experiments.
The bag-level ice probability is computed as the 
mean of the instance predictions over the selected patches:
\(\hat{y}_{i}^{\mathrm{ice}} =
\frac{1}{|\mathcal{T}_i|}
\sum_{j \in \mathcal{T}_i}
p_{ij}^{\mathrm{ice}}\).
The binary module is optimized with a focal loss using only bag-level water--ice 
labels. This module encourages the feature extractor to learn representations that 
distinguish ice-related SAR and multimodal patterns from open water.

\subsection{Inter-module Patch Selection}

Between the ice-water classification module and the ice-type proportion prediction
module, we insert an attention-guided clustering step to select a compact and
informative subset of patches from each bag. This step addresses the fact that
ice-chart polygons can contain many visually redundant patches, making full-bag
training computationally expensive while providing limited additional information.
After the ice-water classification module, each patch is passed
through the trained module to obtain its feature embedding and attention weight.
The patch embeddings are first reduced with PCA and then clustered using
$k$-means with $k=7$. The clusters are unsupervised modes in feature space, not
class assignments; $k=7$ slightly over-segments relative to the six ice classes
to separate distinct textures---including water-like regions, which occur within
ice polygons because the total ice concentration is below 100\%---without
creating sparsely populated clusters.
For each cluster, we compute a cluster relevance score as the mean attention
weight of the patches assigned to that cluster. We then select a fixed percentage
of the bag's patches (the selection budget) and distribute these slots across
clusters in proportion to their relevance scores. Thus,
clusters with higher attention receive more selected patches, while low-attention
clusters contribute fewer or no patches. Within each cluster, selected patches are chosen by their distance to the cluster centroid, with the closest patches retained first. This combines two criteria:
attention identifies discriminative regions, while clustering preserves visual
diversity and avoids selecting only near-duplicate patches. The selected subset is
then used as the input bag for the ice-type proportion prediction module.

\subsection{Sea Ice Type Proportion Prediction Module}
The second module addresses the analyst's fine decision: given that a
region contains ice, what is the concentration of each ice type?
After inter-module patch selection, the sea ice type proportion prediction module
operates on higher-resolution patches from the ice
polygon. The higher-resolution
patches are used here because distinguishing ice stages of development requires
finer textural detail than separating water from ice. The target is a multi-label proportion
vector $\mathbf{y}_i \in \Delta^{K-1}$ where up to
three ice types may be present in the same polygon with corresponding partial
concentrations. This module is trained under a multi-label proportion learning (MLPL) formulation, since each polygon may contain several ice
types simultaneously and the supervision specifies their partial concentrations
rather than a single class label. Each selected patch $x_{ij}$ is encoded by the proportion prediction backbone to obtain a feature vector $\mathbf{h}_{ij} = \phi_{\theta}^{\mathrm{prop}}(x_{ij})$, $\mathbf{h}_{ij} \in \mathbb{R}^{D}$,
where $\phi_{\theta}^{\mathrm{prop}}$ denotes the backbone of the proportion
prediction module. The backbone consists of a ResNet50 feature extractor followed
by a feature projection layer. A patch-level instance head then produces class
logits $\mathbf{o}_{ij} = r_{\theta}^{\mathrm{prop}}(\mathbf{h}_{ij})$, $\mathbf{o}_{ij} \in \mathbb{R}^{K}$.

For the standard proportion model, the logits are mapped to patch-level class
proportions using a simplex-valued activation, $\mathbf{p}_{ij}=A(\mathbf{o}_{ij})$, where $A(\cdot)$ is either softmax or, in the sparse variant,
$\mathrm{entmax}_{1.5}$. The entmax activation allows sparse class-proportion
vectors, which is appropriate for ice-chart polygons because only a small subset
of the possible ice classes is usually present. The bag-level proportion estimate
is obtained by averaging the patch-level proportion vectors over the selected bag:
\begin{equation}
    \hat{\mathbf{y}}_i
    =
    \frac{1}{N_i}
    \sum_{j=1}^{N_i}
    \mathbf{p}_{ij}.
    \label{eq:softmax-mean}
\end{equation}
This softmax-mean aggregation defines the baseline proportion model. The model is
optimized by matching this bag-level prediction to the egg-code-derived
concentration vector:
\begin{equation}
\mathcal{L}_{\mathrm{prop}}
=
D_{\mathrm{KL}}
\left(
\mathbf{y}_i
\,\|\,
\hat{\mathbf{y}}_i
\right).
\label{eq:kl-prop}
\end{equation}

\paragraph{Evidential Dirichlet Proportion Head.}
The softmax-mean baseline maps each patch to a normalized class-proportion
vector and then averages these vectors across the bag. This imposes a
sum-to-one constraint on every patch prediction, so each patch contributes the
same total probability mass regardless of its informativeness. In weakly
supervised sea-ice proportion prediction, this assumption can be restrictive:
ambiguous or texture-poor patches are averaged with highly informative patches
and may dilute the bag-level prediction toward a less discriminative
composition.

To relax this constraint, we use an evidential Dirichlet proportion head inspired
by evidential deep learning~\cite{sensoy2018evidential}. Rather
than applying softmax to the patch logits, the model maps each patch logit vector
$\mathbf{o}_{ij}$ to a non-negative evidence vector
$\mathbf{e}_{ij}=\mathrm{softplus}(\mathbf{o}_{ij})
=\log(1+\exp(\mathbf{o}_{ij}))$. Unlike softmax probabilities, the evidence
values are not constrained to sum to one. Their relative magnitudes determine
the class direction of the prediction, while their absolute magnitudes allow
patches to contribute different amounts of evidence.

For polygon $i$, patch-level evidence is aggregated across the bag as
$\mathbf{E}_{i}=\sum_{j=1}^{N_i}\mathbf{e}_{ij}$. Because polygons contain
different numbers of patches, the aggregated evidence is normalized by bag size
before being converted into a Dirichlet concentration parameter,
$\boldsymbol{\alpha}_{i}=\mathbf{1}+s\mathbf{E}_{i}/N_i$, where $s$ is an
evidence scaling factor and $\mathbf{1}$ denotes a uniform Dirichlet prior. The
predicted bag-level proportion vector is the mean of the Dirichlet distribution,
$\hat{\mathbf{y}}_i=\boldsymbol{\alpha}_{i}/\alpha_{i0}$, where
$\alpha_{i0}=\sum_{c=1}^{K}\alpha_{ic}$. Thus, the direction of
$\boldsymbol{\alpha}_{i}$ determines the predicted ice-type composition, while
the total concentration $\alpha_{i0}$ provides a relative proxy for the amount of
evidence accumulated for the bag. In this work, we use the Dirichlet formulation
primarily as an evidence-weighted aggregation mechanism, rather than as a fully
calibrated uncertainty estimator.

The target polygon-level proportion vector is also represented as a Dirichlet
distribution. To avoid degenerate zero entries, the target vector is first
Laplace-smoothed as
$\tilde{\mathbf{y}}_i=(\mathbf{y}_i+\varepsilon/K)/(1+\varepsilon)$. The
smoothed target is then converted into a target Dirichlet parameter
$\boldsymbol{\beta}_i=\mathbf{1}+\kappa\tilde{\mathbf{y}}_i$, where $\kappa$
controls the concentration of the target distribution. The evidential head is
trained by minimizing the closed-form KL divergence between the predicted and
target Dirichlet distributions,
$\mathcal{L}_{\mathrm{Dir}}=
D_{\mathrm{KL}}\!\left[\mathrm{Dir}(\boldsymbol{\alpha}_i)\,\|\,
\mathrm{Dir}(\boldsymbol{\beta}_i)\right]$. Unlike the simplex KL of the
softmax-mean baseline in Eq.~\eqref{eq:kl-prop}, which places the target first,
the closed-form Dirichlet divergence places the predicted distribution
$\mathrm{Dir}(\boldsymbol{\alpha}_i)$ as its first argument; this is the standard
form of $D_{\mathrm{KL}}$ between two Dirichlet distributions and matches our
implementation. Averaged over the $K$ classes for scale comparability with the
baseline loss, this divergence expands as
\begin{align}
    \mathcal{L}_{\mathrm{Dir}}
    =
    \frac{1}{K}
    \Bigg(
    &\log\Gamma(\alpha_{i0})
    -
    \log\Gamma(\beta_{i0})
    \notag\\
    &-
    \sum_{c=1}^{K}
    \left[
        \log\Gamma(\alpha_{ic})
        -
        \log\Gamma(\beta_{ic})
    \right]
    \notag\\
    &+
    \sum_{c=1}^{K}
    (\alpha_{ic}-\beta_{ic})
    \left[
        \psi(\alpha_{ic})
        -
        \psi(\alpha_{i0})
    \right]
    \Bigg),
    \label{eq:dirichlet-kl}
\end{align}
where $\Gamma(\cdot)$ and $\psi(\cdot)$ denote the gamma and digamma functions,
respectively. The $1/K$ factor is a constant rescaling that leaves the optimum
unchanged and can equivalently be absorbed into $\lambda_{\mathrm{Dir}}$. For the
Dirichlet variant, this term constitutes the proportion objective,
$\mathcal{L}_{\mathrm{prop}}=\lambda_{\mathrm{Dir}}\mathcal{L}_{\mathrm{Dir}}$,
which replaces Eq.~\eqref{eq:kl-prop}.

This objective encourages the predicted Dirichlet mean to match the polygon-level
ice-type proportions while also regularizing the total evidence scale. A
prediction with an incorrect class composition and large concentration is
penalized more strongly than a similarly incorrect but low-evidence prediction.
However, since no patch-level confidence or uncertainty labels are available, we
interpret $\alpha_{i0}$ only as a relative confidence proxy. Calibration of this
quantity is evaluated separately and is not assumed by the model formulation.


\subsection{Multi-Modal Extensions}

The proportion prediction module described above operates exclusively on SAR
imagery. While SAR captures the textural and geometric signatures of ice
surfaces, the ice type present in a polygon is also influenced by thermodynamic
and oceanographic conditions that a SAR snapshot alone cannot observe. We extend the
proportion prediction module to incorporate three complementary sources of
environmental context: AMSR2 passive microwave brightness temperatures, ERA5
atmospheric reanalysis fields, and auxiliary geometric scalars. These are fused
with the SAR backbone in one of two ways, depending on how the non-SAR channels
are coupled with the spatial feature extractor.

Each patch is represented as a 16-channel tensor. The channels are organized as
follows: three SAR channels (HH, HV, and incidence angle), one distance-to-land channel, four AMSR2 brightness
temperature channels ($T_{18.7}^{H}$, $T_{18.7}^{V}$, $T_{36.5}^{H}$,
$T_{36.5}^{V}$), five ERA5 channels (rotated 10\,m u and v wind components,
2\,m temperature, total column water vapor, and total column cloud liquid
water), two latitude/longitude channels, and one normalized month channel.
These inputs provide complementary information: SAR captures local backscatter texture and structure at patch resolution, while AMSR2, ERA5, geolocation, and month provide broader thermodynamic, atmospheric, and geographic context. Because these non-SAR variables are generally smoother and less spatially resolved than SAR within an individual patch, we evaluate two fusion strategies.

\paragraph{Early Fusion}

In the early-fusion variant, all 16 channels are passed jointly through the
ResNet50 backbone. The first convolutional layer is adapted to accept 16 input
channels, with ImageNet weights retained for the three SAR channels and the
remaining thirteen initialized to zero.

\paragraph{Late Fusion}

In the late-fusion variant, SAR channels are processed by the ResNet50 backbone
while the thirteen environmental channels are spatially averaged and encoded
by a shallow MLP, producing $\mathbf{h}_{ij}^{\mathrm{env}} \in
\mathbb{R}^{D_{\mathrm{env}}}$. The two feature vectors are concatenated before
the instance proportion head. Since AMSR2, ERA5, and auxiliary scalars are
spatially quasi-constant at patch resolution, spatial averaging is appropriate
and avoids applying convolutional processing to uninformative spatial structure.

\paragraph{Modality-Guided Auxiliary Regularization}
Building on the late-fusion architecture, this variant adds two training-time
auxiliary regularizers while leaving the inference architecture unchanged. Both
regularizers are shallow MLPs trained end-to-end with the proportion head: one
encourages consistency between the predicted total ice concentration and an
AMSR2-derived ice-concentration signal, and the other provides an ERA5-driven
environmental-compatibility prior.

The first auxiliary loss encourages the predicted total ice concentration to be
consistent with a signal learned from AMSR2 passive-microwave brightness
temperatures. A three-layer MLP \(g_{\theta}^{\mathrm{SIC}}\) maps the bag-mean
AMSR2 vector \(\bar{\mathbf{s}}_{i}^{\mathrm{AMSR2}} \in \mathbb{R}^{4}\),
formed from \((18.7\mathrm{H}, 18.7\mathrm{V}, 36.5\mathrm{H},
36.5\mathrm{V})\), to a scalar ice fraction \(\hat{c}_i \in [0,1]\). The MLP
uses hidden layers of width 32 and 16 with ReLU activations and a final sigmoid.
The proportion head is encouraged to predict an ice total, defined as one minus
the water-class proportion, that matches this estimate:
\begin{equation}
\mathcal{L}_{\mathrm{SIC}}
=
\left(
\sum_{c \neq \mathrm{water}} \hat{y}_{ic}
-
g_{\theta}^{\mathrm{SIC}}
\left(
\bar{\mathbf{s}}_{i}^{\mathrm{AMSR2}}
\right)
\right)^2 .
\end{equation}
We learn the AMSR2-derived ice-concentration estimate directly from the bag-mean
brightness temperatures, coupling the SAR-driven proportion estimate to a
microwave-derived total-ice signal.                                                                                                                                                                      

The second auxiliary loss, \(\mathcal{L}_{\mathrm{env}}\), is a learned
environmental-compatibility regularizer. Ice type reflects thickness and age
accumulated over weeks to years and is interpreted primarily from SAR
backscatter and texture, supplemented by passive-microwave and optical/infrared
imagery~\cite{manice, lavergne2024panarctic}. Atmospheric state does not
determine ice type, but it correlates with ice growth and melt regimes, so ERA5
is used only as a soft, data-driven prior over plausible ice types.
A two-layer MLP \(g_{\theta}^{\mathrm{env}}\) maps the bag-mean ERA5 vector
\(\bar{\mathbf{e}}_{i} \in \mathbb{R}^{5}\), consisting of rotated
\(10\,\mathrm{m}\) \(u\) and \(v\) wind components, \(2\,\mathrm{m}\)
temperature, total column water vapor, and total column cloud liquid water, to
a per-class compatibility vector \(\boldsymbol{\pi}_i \in [0,1]^K\). The MLP
uses one hidden layer of width 32 with ReLU activation and a final sigmoid. Each
entry \(\pi_{ic}\) is an end-to-end learned compatibility score. The regularizer discourages the
proportion head from assigning mass to classes with low learned compatibility:
\begin{equation}
\mathcal{L}_{\mathrm{env}}(i)
=
\sum_{c=1}^{K}
\hat{y}_{ic}
\bigl(
1 - \pi_{ic}
\bigr).
\end{equation}

The final training objective is
\begin{equation}
\mathcal{L}
=
\mathcal{L}_{\mathrm{prop}}
+
\lambda_{\mathrm{SIC}}\mathcal{L}_{\mathrm{SIC}}
+
\lambda_{\mathrm{env}}\mathcal{L}_{\mathrm{env}},
\end{equation}
where \(\lambda_{\mathrm{SIC}}\) and \(\lambda_{\mathrm{env}}\) control the
relative contribution of the two auxiliary terms. Both auxiliary heads are trained end-to-end from the polygon-level objective, with the environmental loss acting as a soft regularizer whose influence decreases as the learned compatibility scores increase. Its contribution is further limited by $\lambda_{\mathrm{env}} = 0.1$, with $\lambda_{\mathrm{SIC}} = 0.5$. Both auxiliary losses are used
only during training; at inference, predictions are produced by the proportion
head alone.

\section{Experimental Evaluation}
This section describes the dataset, preprocessing pipeline, model configurations, training setup, evaluation metrics, baselines, and experimental results used to assess the proposed framework.
\subsection{Dataset and Preprocessing}
We evaluate on the raw AI4Arctic Sea Ice Challenge Dataset~\cite{ChallengeDataset, autoice}, released by DTU, DMI, and NERSC for the ESA AutoICE challenge. The dataset contains 513 Sentinel-1 Extra-Wide GRD scenes acquired over the Canadian and Greenlandic Arctic between January 2018 and December 2021. Each scene has $40\,\text{m}$ pixel spacing, dual-polarization HH/HV SAR channels, co-located AMSR2 passive-microwave measurements, ERA5 weather variables, and manually produced WMO egg-code ice charts from the Greenland Ice Service and Canadian Ice Service.
We use the raw dataset because it preserves the complete polygon-level egg-code information, including all ice types and their partial concentrations. This proportion-based supervision is required by the proposed model formulation. WMO stage-of-development codes are grouped into six ice-type classes, listed in Table~\ref{tab:wmo_mapping}; an additional open-water class $C_6$ is
recovered within each ice polygon as the residual $1-CT$.
\begin{table}[!t]
\centering
\caption{Mapping from WMO stage-of-development codes to target classes.}
\label{tab:wmo_mapping}
\scriptsize
\setlength{\tabcolsep}{2pt}
\renewcommand{\arraystretch}{0.95}
\begin{tabular}{lll}
\toprule
\textbf{ID} & \textbf{Class} & \textbf{WMO codes} \\
\midrule
C0 & New/Nilas & 81, 82 \\
C1 & Young/Grey/White & 83, 84, 85 \\
C2 & Thin First-Year & 87, 88, 89 \\
C3 & Med./Thick First-Year & 86, 91, 93 \\
C4 & Old/Multi-Year & 95, 96, 97 \\
C5 & Glacier & 98 \\
\bottomrule
\end{tabular}
\end{table}

\begin{table}[!t]
\centering
\caption{Input channels for SAR-only and multimodal settings.}
\label{tab:input_channels}
\scriptsize
\setlength{\tabcolsep}{2pt}
\renewcommand{\arraystretch}{0.95}
\begin{tabular}{p{0.30\columnwidth}p{0.62\columnwidth}}
\toprule
\textbf{Group} & \textbf{Definition} \\
\midrule
SAR + inc., 3 ch. &
Sentinel-1 HH and HV backscatter and incidence angle. \\

Dist.-to-land, 1 ch. &
Distance to nearest land/coastline. \\

AMSR2, 4 ch. &
Brightness temperatures at 18.7 and 36.5 GHz, H/V polarizations. \\

ERA5, 5 ch. &
Near-surface wind, temperature, and atmospheric water variables. \\

Geo. + month, 3 ch. &
Latitude, longitude, and normalized acquisition month. \\
\midrule
SAR-only &
3 channels: Sentinel-1 HH and HV + incidence. \\

Multimodal &
16 channels: all groups above. \\
\bottomrule
\end{tabular}
\end{table}
Each ice-chart polygon is rasterized onto the SAR grid and tiled into non-overlapping $20 \times 20$-pixel patches. Before patch extraction, 203 polygons with inconsistent concentration labels were removed, retaining only those satisfying \(C_T = C_A + C_B + C_C\). Candidate patches are kept when at least 10\% of their pixels lie inside the polygon; remaining out-of-polygon pixels are filled using nearest-neighbour imputation from in-polygon pixels.
Module~2 uses patches at the raw SAR resolution ($40\,\mathrm{m}$), while Module~1 uses a coarser patch set obtained by downsampling the SAR grid by a factor of 5 ($200\,\mathrm{m}$). Auxiliary non-SAR channels are resampled using bilinear interpolation to the SAR grid. Patch locations are defined at the raw grid-spacing and projected to the coarser grid to preserve alignment. Before inter-module patch selection, bags contained 1--25{,}414 raw-resolution patches, with a mean of 2{,}973.4 patches per bag.

The input channels for each setting are summarised in Table~\ref{tab:input_channels}. To ensure reliable bag-level estimates, we exclude polygons with fewer than 1 patch for Module~1 and fewer than 10 patches for Module~2. All channels are $z$-score normalized using training-set statistics, and bags are randomly split into 70\,/\,15\,/\,15 train, validation, and test sets. To assess possible scene-level overlap under this random polygon split, we additionally evaluate a scene-disjoint split in which all polygons from the same Sentinel-1 scene are assigned to a single subset.
After filtering, Module~1 contains 7,118 bags: 628 water and 6,490 ice, split into 4,984 training, 1,067 validation, and 1,067 test bags. Module~2 contains 6,136 ice bags after removing polygons with fewer than 10 valid patches, split into 4,296 training, 920 validation, and 920 test bags.
\begin{figure}[!t]
  \centering
  \includegraphics[width=\columnwidth]{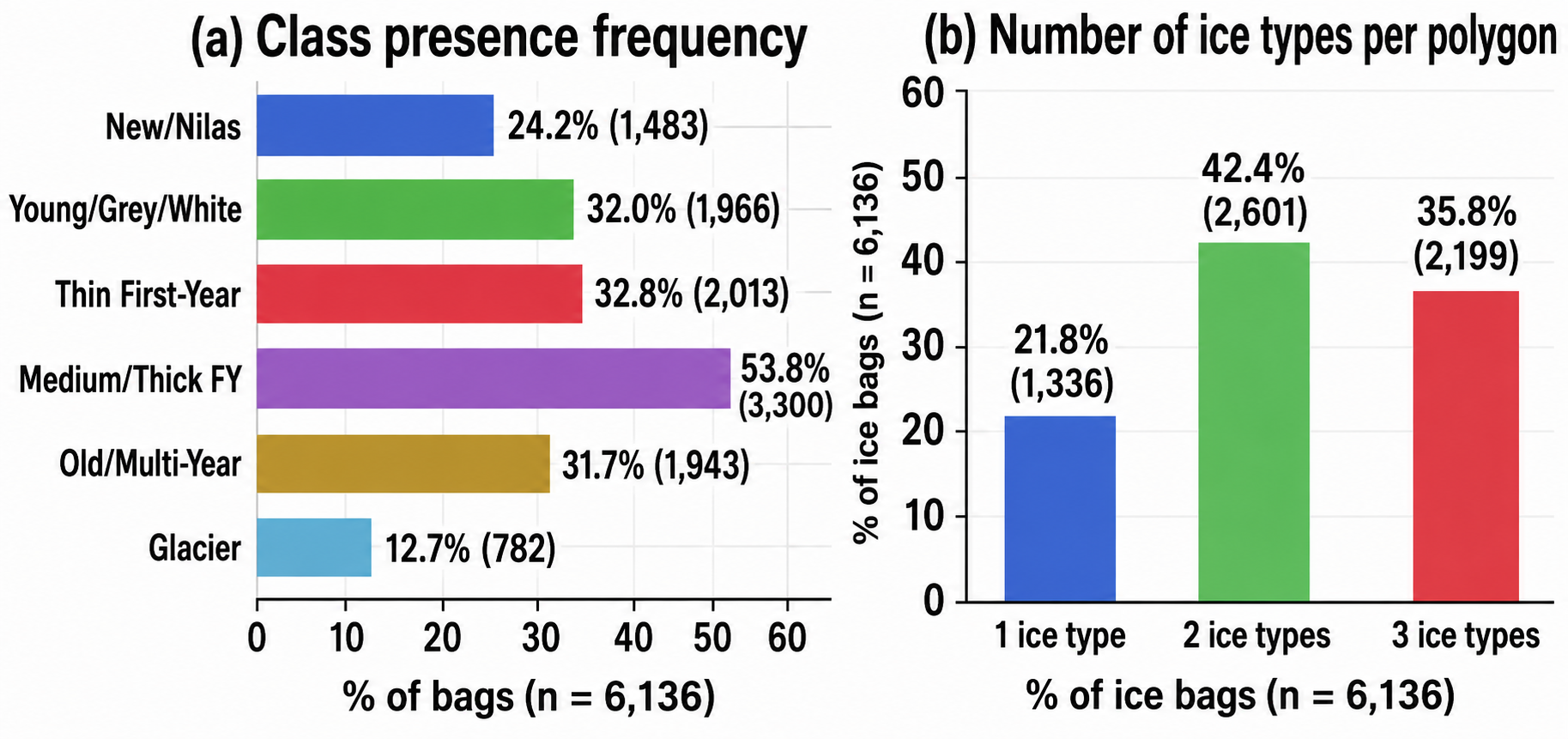}
  \caption{Dataset label statistics.
           \textbf{(a)}~Class presence frequency across ice classes.
           \textbf{(b)}~Distribution of the number of ice types per polygon.}
           \Description{Two bar charts. Left: class presence frequency across six ice classes, with Medium/Thick First-Year most frequent at 53.8 percent of bags and Glacier rarest at 12.7 percent. Right: distribution of ice types per polygon, with 21.8 percent of polygons containing one ice type, 42.4 percent containing two, and 35.8 percent containing three.}
  \label{fig:label_dist}
  \vspace{-4mm}
\end{figure}

\begin{figure}[!t]
  \centering
  \includegraphics[width=\columnwidth]{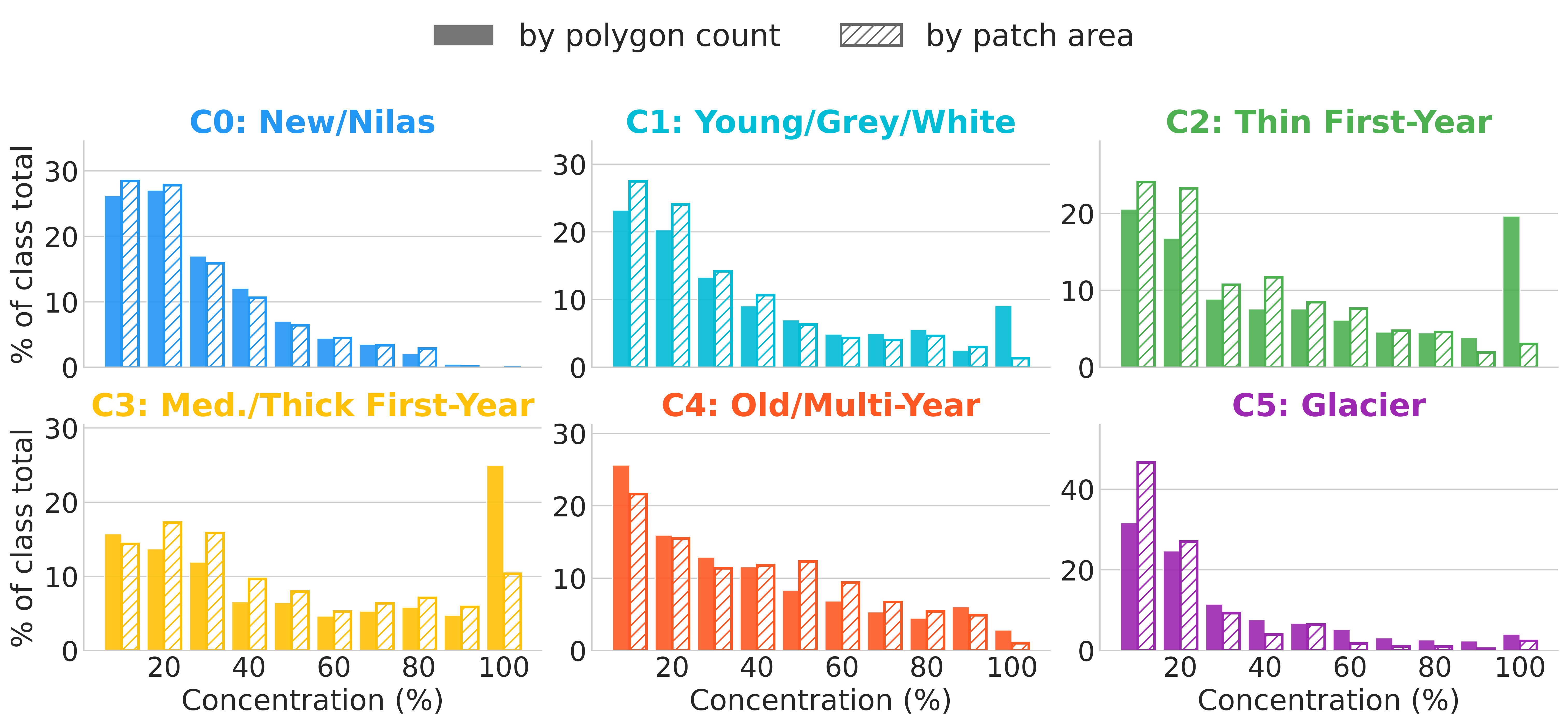}
  \caption{Per-class concentration distributions by polygon count (solid) and by patch area (hatched).}
  \Description{Six histograms, one per ice class, showing the distribution of partial concentration values from 10 to 100 percent, with solid bars weighted by polygon count and hatched bars weighted by patch area. Most classes concentrate at low values of 10 to 20 percent, while Medium/Thick First-Year also shows a peak at 100 percent concentration.}
  \label{fig:conc_dist}
\end{figure}

Figure~\ref{fig:label_dist} summarizes the label statistics of the ice-chart polygon dataset. Medium/Thick FY is the most frequent class, appearing in 53.8\% of bags, while Glacier (generally icebergs originating from land-ice) is the rarest at 12.7\%. Most polygons are multi-label: 42.4\% contain two ice types and 35.8\% contain three, meaning 78.2\% contain more than one ice type.
Figure~\ref{fig:conc_dist} shows per-class concentration distributions conditioned on class presence, by polygon count (solid) and patch area (hatched).  Medium/Thick FY is the most common class ($53.8\%$ of polygons) and Glacier the rarest ($12.7\%$); $78.2\%$ of polygons are multi-label.
Patch-area weighting shifts these: Glacier covers just $2.6\%$ of ice area, while Old/Multi-Year rises from $31.7\%$ to $47.6\%$.
To address imbalance in the SAR-only model, we use three sampling strategies: undersampling bags dominated by a frequent class with no rare class present, ensuring each rare class appears at least once per mini-batch, and repeating bags with $\geq 60\%$ concentration of a target class $3\times$. The multimodal model uses the unchanged training set so that the effect of auxiliary channels can be evaluated without sampling interventions.

\subsection{Baselines }
To evaluate the proposed framework, we compare against supervised IceBench~\cite{alkaeetaleghan2025icebench} patch classifiers (CNN, AlexNet, VGG-16, DenseNet-121, and ResNet-50/101/152). Each patch is assigned the dominant ice type of its parent polygon ($\arg\max_c y_c$) and trained with weighted cross-entropy; a WeightedRandomSampler further balances training samples. Bag-level proportions are obtained by averaging patch-level softmax predictions.  
For weakly supervised baselines, we adapt DeepLLP~\cite{ardehaly2017}, ABMIL~\cite{Ilse2018}, and TransMIL~\cite{Shao2021TransMIL} to the same polygon-level proportion setting. In our implementation, DeepLLP uses a ResNet-50 patch encoder followed by a softmax instance-proportion head, and bag proportions are obtained by averaging patch-level predictions. ABMIL replaces the uniform averaging of patch-level predictions with learned attention-based aggregation; in our experiments, we evaluate its class-specific attention variant. TransMIL is adapted as a context encoder: patch embeddings are first contextualized with Transformer self-attention, and the resulting patch-level proportions are then aggregated and trained against the polygon-level concentration vector. All weakly supervised baselines use the same selected bags and are optimized using KL-based proportion supervision. We additionally evaluate two internal LLP variants. Softmax + WKL uses a single-head proportion model with softmax-normalized patch predictions, bag-level averaging, and weighted KL divergence to address class imbalance in the concentration targets. Entmax + asym.(asymmetric) replaces softmax with \(\mathrm{entmax}_{1.5}\) to obtain sparser patch-level distributions and adds an asymmetric penalty to reduce underprediction of rare or high-concentration ice classes. Softmax LLP and Entmax LLP are internal LLP variants included as controlled weakly-supervised comparisons.

\subsection{Experimental Setup}
Both modules use separate ResNet-50 backbones with 512d feature
projections and dropout $0.2$. Module~1 is trained as an attention-top-$k$ MIL
classifier with $k=10$, attention temperature $\tau=0.5$, focal loss
($\gamma=2$, $\alpha=0.25$), Adam optimizer, learning rate
$5{\times}10^{-5}$, weight decay $10^{-5}$, and early stopping patience 15,
for up to 100 epochs. Module~2 is trained on the top 25\% attention-selected patches. For the SAR-only Dirichlet model, module~2 is
trained for up to 200 epochs using Adam with learning rate $5{\times}10^{-5}$,
weight decay $10^{-5}$, cosine scheduling, and early stopping patience 10. The
Dirichlet head uses mean-scale evidence aggregation with $s=\kappa=100$,
$\varepsilon=0.01$, and $\lambda_{\mathrm{Dir}}=1$. For the multimodal models, module~2 is trained for up to 100 epochs using Adam
with learning rate $10^{-4}$, weight decay $10^{-5}$, cosine scheduling, and
early stopping patience 15. All experiments were run on an NVIDIA RTX PRO 6000 GPU.

\subsection{Evaluation Metrics}
For evaluation metric, the ice-water classification module is evaluated as a binary bag-level
classification problem. We report bag-level accuracy, precision, recall,
F1 score, and AUROC, with Ice treated as the positive class.
For the sea ice type proportion prediction module, the primary metric is
per-class mean absolute error (MAE) between the predicted and true bag-level
proportions: 
\begin{equation}
  \mathrm{MAE}_c
    =
    \frac{1}{|\mathcal{B}_c|}
    \sum_{i \in \mathcal{B}_c}
    \left|
    y_{ic} - \hat{y}_{ic}
    \right|.
  \label{eq:mae}
\end{equation}
Here, $\mathcal{B}_c$ denotes the set of test bags in which class $c$ is present.
The overall MAE is reported as the unweighted mean of the per-class MAE values
across the ice classes. Presence detection is evaluated by thresholding predicted proportions,
\(\hat{z}_{ic}=\mathbb{1}[\hat{y}_{ic}>0.1]\), using chart-derived labels
\(z_{ic}\) as ground truth. We report class-wise binary F1 scores; the reported average (Avg) is
unweighted average across ice classes.


\subsection{Results}
Table~\ref{tab:stage1_results} reports module~1 test metrics across both
configurations. The SAR-only attention-topk model achieves an F1
score of 94.7 and AUC of 96.1, establishing a strong binary detection
baseline.  Its high precision (99.0) shows that ice predictions are rarely false
positives, while recall (90.7) indicates that only a small fraction of ice
polygons are missed. The 16-channel multimodal early-fusion model performs worse, with F1=90.8 and
AUC=87.9. This likely reflects a scale mismatch: the auxiliary AMSR2, ERA5,
geolocation, and seasonal channels are coarse and have similar or the same values for large groups of neighboring patches,
so they add limited information for this binary task and may dilute the SAR
signal.

\begin{table}[!t]
\centering
\caption{Module~1 test results (binary water vs.\ ice).}
\label{tab:stage1_results}
\small
\setlength{\tabcolsep}{4pt}
\begin{tabular}{lcccccc}
\toprule
\textbf{Configuration} & \textbf{Ch.} & \textbf{Acc.} & \textbf{Prec.} & \textbf{Rec.} & \textbf{F1} & \textbf{AUC} \\
\midrule
SAR-only  & 3  & 90.6 & 99.0 & 90.7 & 94.7 & 96.1 \\
Multimodal Early Fusion & 16 & 84.4 & 97.6 & 84.9 & 90.8 & 87.9 \\
\bottomrule
\end{tabular}
\end{table}

Table~\ref{tab:mae_sar} reports per-class proportion MAE and Table~\ref{tab:f1_sar} reports per-class presence F1 for SAR-only input. Supervised IceBench classifiers perform poorly because dominant-class patch labels do not capture the mixed-proportion structure of ice-chart polygons. DenseNet-121 is the strongest supervised baseline, but it still falls well behind all weakly supervised proportion-learning methods. Proportion-learning baselines improve substantially by optimizing directly on polygon-level concentration vectors. DeepLLP lowers MAE to 0.275 and raises F1 to 40.0, while ABMIL performs worst among the proportion-learning baselines, suggesting that shared attention pooling is insufficient for multi-label ice mixtures. TransMIL improves further through inter-patch context, reaching 0.267 MAE and 47.6 F1. The LLP internal variants that built on the same backbone, bags, and KL supervision as our model perform best among non-Dirichlet methods: WKL and Entmax raise F1 to 52.0 and 54.2. However, both retain higher $C_3$ MAE than the proposed model, indicating residual bias in high-concentration proportion prediction. 
The proposed Dirichlet head achieves the best overall performance, with 0.247 MAE and 54.8 F1, outperforming all baselines on both aggregate metrics. On a per-class basis it attains the lowest MAE for $C_2$, $C_3$, and $C_6$ and the highest F1 for every class except $C_5$. Its largest MAE reductions over Entmax occur on $C_3$ (0.387$\rightarrow$0.320), $C_2$ (0.308$\rightarrow$0.276), and $C_6$ (0.200$\rightarrow$0.168). By using non-negative patch evidence instead of a strict softmax constraint, the model down-weights ambiguous patches and reduces dilution from boundary or texture-poor regions.

We also evaluate two instance-label-reconstruction LLP methods,
EasyLLP~\cite{busafekete2023} and LLP-BP~\cite{havaldar2024llpbp},
adapted to the same bag structure. EasyLLP produced a proportion MAE of 0.372, while LLP-BP improved to 0.268. However, both methods still underperformed the Dirichlet head. This suggests that methods aimed at reconstructing instance labels from bag proportions are less suitable for ambiguous sea-ice patches, whereas the Dirichlet head better models patch contributions as uncertain evidence.

\begin{table}[t]
  \centering
  \caption{Module~2 per-class proportion MAE --- \textbf{SAR-only}.}
  \label{tab:mae_sar}
  \setlength{\tabcolsep}{1.5pt}
  \renewcommand{\arraystretch}{1.05}
  \scriptsize
  \begin{tabular}{lcccccccc}
    \toprule
    Method & $C_0$ & $C_1$ & $C_2$ & $C_3$ & $C_4$ & $C_5$ & $C_6$ & All \\
    \midrule
    \multicolumn{9}{l}{\emph{Supervised classifiers (IceBench)}} \\
    CNN          & 0.278 & 0.319 & 0.386 & 0.530 & 0.265 & 0.387 & 0.353 & 0.361 \\
    AlexNet      & 0.302 & 0.338 & 0.397 & 0.530 & 0.277 & 0.362 & 0.366 & 0.368 \\
    VGG-16       & 0.235 & 0.320 & 0.350 & 0.514 & 0.233 & 0.331 & 0.284 & 0.331 \\
    DenseNet-121 & 0.229 & 0.285 & 0.285 & 0.493 & 0.207 & 0.233 & 0.294 & 0.289 \\
    ResNet-50    & 0.255 & 0.284 & 0.359 & 0.494 & 0.217 & 0.288 & 0.322 & 0.316 \\
    ResNet-101   & 0.264 & 0.314 & 0.378 & 0.529 & 0.245 & 0.362 & 0.341 & 0.349 \\
    ResNet-152   & 0.235 & 0.301 & 0.385 & 0.515 & 0.250 & 0.376 & 0.366 & 0.344 \\
    \midrule
    \multicolumn{9}{l}{\emph{Weakly-supervised proportion-learning baselines}} \\
    DeepLLP             & 0.232 & 0.249 & 0.302 & 0.328 & 0.256 & 0.283 & 0.299 & 0.275 \\
    ABMIL     & 0.221 & 0.241 & 0.341 & 0.342 & 0.247 & 0.322 & 0.287 & 0.286 \\
    TransMIL            & 0.194 & 0.222 & 0.315 & 0.391 & 0.253 & \textbf{0.226} & 0.266 & 0.267 \\
    Softmax + WKL           & 0.173 & 0.222 & 0.323 & 0.342 & \textbf{0.198} & 0.273 & 0.288 & 0.255 \\
    Entmax + asym           & \textbf{0.168} & \textbf{0.218} & 0.308 & 0.387 & 0.202 & 0.237 & 0.200 & 0.253 \\
    \midrule
    \multicolumn{9}{l}{\emph{Proposed model }} \\

    \rowcolor{proprow}
    Dirichlet & 0.184 & 0.235 & \textbf{0.276} & \textbf{0.320} & 0.207 & 0.260 & \textbf{0.168} & \textbf{0.247} \\
    \bottomrule
  \end{tabular}
\end{table}
\begin{table}[t]
  \centering
  \caption{Module~2 per-class presence F1 --- \textbf{SAR-only}.}
  \label{tab:f1_sar}
  \setlength{\tabcolsep}{1.5pt}
  \renewcommand{\arraystretch}{1.05}
  \scriptsize
  \begin{tabular}{lcccccccc}
    \toprule
    Method & $C_0$ & $C_1$ & $C_2$ & $C_3$ & $C_4$ & $C_5$ & $C_6$ & Avg \\
    \midrule
    \multicolumn{9}{l}{\emph{Supervised classifiers (IceBench)}} \\
    CNN          & 15.5 & 11.9 & 10.3 & 0.0 & 25.3 & 11.3 & 12.0 & 12.4 \\
    AlexNet      & 15.9 & 3.5 & 10.1 & 0.0 & 17.7 & 10.7 & 4.7 & 9.7 \\
    VGG-16       & 16.6 & 8.6 & 16.6 & 0.0 & 34.1 & 13.1 & 34.5 & 14.8 \\
    DenseNet-121 & 17.9 & 12.0 & 39.1 & 3.4 & 50.2 & 20.6 & 31.6 & 23.9 \\
    ResNet-50    & 16.0 & 14.9 & 25.0 & 4.7 & 42.8 & 14.3 & 16.5 & 19.6 \\
    ResNet-101   & 16.6 & 11.2 & 10.9 & 0.0 & 32.9 & 11.7 & 16.6 & 13.9 \\
    ResNet-152   & 16.2 & 8.8 & 14.8 & 0.0 & 39.6 & 10.7 & 7.5 & 15.0 \\
    \midrule
    \multicolumn{9}{l}{\emph{Weakly-supervised proportion-learning baselines}} \\
    DeepLLP             & 10.7 & 48.7 & 47.7 & 71.2 & 47.1 & 14.6 & 47.5 & 40.0 \\
    ABMIL    & 5.4 & 48.5 & 43.4 & 71.6 & 49.7 & 0.0 & 52.9 & 36.4 \\
    TransMIL            & 28.0 & 48.4 & 47.6 & 71.6 & 55.5 & 34.6 & 66.2 & 47.6 \\
    Softmax + WKL            & 44.1 & 50.2 & 54.0 & 71.0 & 51.9 & 40.7 & 55.7 & 52.0 \\
    Entmax + asym           & 42.8 & 52.2 & 52.2 & 68.8 & 59.6 & \textbf{49.8} & 73.3 & 54.2 \\
    \midrule
    \multicolumn{9}{l}{\emph{Proposed Model}} \\

    \rowcolor{proprow}
    Dirichlet  & \textbf{46.4} & \textbf{54.1} & \textbf{54.6} & \textbf{73.4} & \textbf{62.8} & 37.3 & \textbf{76.6} & \textbf{54.8} \\
    \bottomrule
  \end{tabular}
\end{table}

\begin{table}[t]
  \centering
  \caption{Module~2 per-class proportion MAE --- \textbf{Multimodal}.}
  \label{tab:mae_mm}
  \setlength{\tabcolsep}{1.5pt}
  \renewcommand{\arraystretch}{1.05}
  \scriptsize
  \begin{tabular}{lcccccccc}
    \toprule
    Method & $C_0$ & $C_1$ & $C_2$ & $C_3$ & $C_4$ & $C_5$ & $C_6$ & All \\
    \midrule
    \multicolumn{9}{l}{\emph{Supervised classifier}} \\
    DenseNet-121  & 0.300 & 0.321 & 0.269 & 0.279 & 0.333 & 0.340 & 0.283 & 0.307 \\
    \midrule
    \multicolumn{9}{l}{\emph{Proposed Model (multimodal)}} \\
    \rowcolor{proprow}
    Early Fusion            & 0.208 & 0.224 & 0.276 & 0.176 & 0.216 & 0.233 & 0.228 & 0.222 \\
    \rowcolor{proprow}
    Late Fusion             & 0.199 & 0.225 & \textbf{0.202} & \textbf{0.168} & 0.224 & \textbf{0.182} & 0.204 & 0.200 \\
    \rowcolor{proprow}
    Late Fusion + Aux-Cons          & \textbf{0.187} & \textbf{0.213} & 0.206 & 0.172 & \textbf{0.204} & \textbf{0.182} & \textbf{0.203} & \textbf{0.194} \\
    \bottomrule
  \end{tabular}
\end{table}

\begin{table}[t]
  \centering
  \caption{Module~2 per-class presence F1 --- \textbf{Multimodal}.}
  \label{tab:f1_mm}
  \setlength{\tabcolsep}{1.5pt}
  \renewcommand{\arraystretch}{1.05}
  \scriptsize
  \begin{tabular}{lcccccccc}
    \toprule
    Method & $C_0$ & $C_1$ & $C_2$ & $C_3$ & $C_4$ & $C_5$ & $C_6$ & Avg \\
    \midrule
    \multicolumn{9}{l}{\emph{Supervised classifier}} \\
    DenseNet-121  & 30.0 & 44.1 & 57.8 & 69.0 & 55.5 & 48.0 & 53.5 & 50.7 \\
    \midrule
    \multicolumn{9}{l}{\emph{Proposed model (multimodal)}} \\
    \rowcolor{proprow}
    Early Fusion            & 70.5 & 72.1 & 70.1 & 85.7 & 82.3 & 60.0 & 73.2 & 73.4 \\
    \rowcolor{proprow}
    Late Fusion             & 77.4 & 74.3 & 72.4 & 85.0 & \textbf{87.6} & \textbf{64.6} & 77.1 & 76.9 \\
    \rowcolor{proprow}
    Late Fusion + Aux-Cons          & \textbf{77.9} & \textbf{75.1} & \textbf{74.8} & \textbf{88.4} & 85.0 & 63.1 & \textbf{78.9} & \textbf{77.4} \\
    \bottomrule
  \end{tabular}
\end{table}
Table~\ref{tab:mae_mm} reports per-class proportion MAE and Table~\ref{tab:f1_mm} reports per-class presence F1 for multimodal input. We denote the late-fusion model with both modality-guided auxiliary regularizers as
LF + Aux-Cons. As the supervised multimodal baseline, we use DenseNet-121, which was selected because it achieved the best performance among the supervised classifiers in the SAR-only comparison. 
DenseNet-121 obtains an overall MAE of $0.307$, while all proposed multimodal variants achieve much lower errors. 
Early Fusion reduces the overall MAE to $0.222$, Late Fusion further improves it to $0.200$, and the LF + Aux-Cons model obtains the best overall MAE of $0.194$.
Late Fusion performs better than Early Fusion, suggesting that SAR and auxiliary environmental variables are more effective when processed separately before feature combination. 
This is likely because AMSR2, ERA5, geolocation, and seasonal variables provide broader contextual information, while SAR carries finer spatial and textural information. 
 The LF + Aux-Cons model gives the lowest MAE for $C_0$, $C_1$, $C_4$, and $C_6$, and matches the best performance for $C_5$, showing that the auxiliary regularizers improve proportion prediction across most classes.
The presence F1 results show a similar trend. 
The supervised DenseNet-121 baseline reaches an average F1 of only $50.7$, while Early Fusion improves this to $73.4$ and Late Fusion to $76.9$. 
The LF + Aux-Cons model achieves the best average F1 of $77.4$.
Overall, auxiliary environmental inputs improve both proportion and presence prediction, with the LF + Aux-Cons model performing best. Late Fusion outperforms Early Fusion by matching each modality to its native scale: AMSR2, ERA5, and geographic scalars are quasi-constant within a patch, so Early Fusion wastes backbone capacity convolving near-flat fields, whereas Late Fusion averages them into a MLP and fuses only at the head, leaving SAR texture intact.

\begin{table}[t]
\centering
\caption{Module-level and end-to-end performance of the proposed framework. }
\label{tab:end_to_end}
\setlength{\tabcolsep}{1.5pt}
\scriptsize
\begin{tabular}{lccccc}
\toprule
\textbf{Framework} &
\textbf{M1 F1} &
\textbf{M2 MAE} &
\textbf{M2 F1} &
\textbf{E2E MAE} &
\textbf{E2E F1} 
 \\
\midrule

SAR-only
& 94.7
& 0.247
& 54.8
& 0.271
& 50.5
 \\

Multimodal
& 90.8
& 0.194
& 77.4
& 0.208
& 69.4
 \\

\bottomrule
\end{tabular}
\end{table}
Table~\ref{tab:end_to_end} compares Module~2 performance evaluated independently with the full end-to-end pipeline. End-to-end composition introduces only a modest increase in MAE, while the reduction in F1 is more noticeable due to errors propagated from Module~1. This shows that most of the end-to-end degradation is caused by gating errors rather than by proportion estimation in Module~2.

\begin{table}[t]
\centering
\caption{Comparison of random polygon-level and scene-disjoint evaluation.}
\label{tab:scene_split}
\setlength{\tabcolsep}{4pt}
\scriptsize
\begin{tabular}{lcccc}
\toprule
\textbf{Method} &
\multicolumn{2}{c}{\textbf{Random split}} &
\multicolumn{2}{c}{\textbf{Scene-disjoint}} \\
\cmidrule(lr){2-3}\cmidrule(lr){4-5}
& \textbf{MAE} & \textbf{F1} & \textbf{MAE} & \textbf{F1} \\
\midrule
Dirichlet  & 0.247 & 54.8 & \textbf{0.260} & \textbf{51.3} \\
Entmax + asym       & 0.253 & 54.2 & 0.265 & 50.0 \\
TransMIL             & 0.267 & 47.6 & 0.273 & 34.6 \\
\bottomrule
\end{tabular}
\end{table}

Table~\ref{tab:scene_split} compares the random polygon-level and
scene-disjoint splits. The proposed Dirichlet model remains the
best-performing method under the stricter scene-disjoint evaluation,
indicating that its advantage is preserved on unseen scenes.

\begin{table}[t]
  \centering
  \caption{Multimodal ablation study (Module~2).}
  \label{tab:ablation_mm}
  \setlength{\tabcolsep}{2pt}
  \renewcommand{\arraystretch}{1.05}
  \scriptsize
  \begin{tabular}{llcccccccc}
    \toprule
    \textbf{Group} & \textbf{Configuration} & $C_0$ & $C_1$ & $C_2$ & $C_3$ & $C_4$ & $C_5$ & $C_6$ & \textbf{All / F1} \\
    \midrule
    \multicolumn{10}{l}{\emph{Auxiliary-regularizer Ablation (Late Fusion Backbone)}} \\
    & No aux. (Late Fusion)     & 0.199 & 0.225 & 0.202 & \textbf{0.168} & 0.224 & \textbf{0.182} & 0.204 & 0.200 / 76.9 \\
    & SIC consistency only          & 0.194 & 0.219 & \textbf{0.198} & 0.170 & 0.212 & 0.189 & 0.205 & 0.197 / 76.5 \\
    & Env. compatibility only       & 0.204 & 0.223 & 0.208 & \textbf{0.168} & 0.207 & 0.183 & \textbf{0.196} & 0.199 / 76.4 \\
    \rowcolor{proprow}
    & Both (LF + Aux-Cons)      & \textbf{0.187} & \textbf{0.213} & 0.206 & 0.172 & \textbf{0.204} & \textbf{0.182} & 0.203 & \textbf{0.194} / \textbf{77.4} \\    
    \bottomrule
  \end{tabular}
\end{table}

\paragraph{Ablation and Sensitivity Analysis}
Table~\ref{tab:ablation_mm} reports the auxiliary-regularizer ablation for
the multimodal late fusion backbone. Removing both auxiliary regularizers already achieves a strong MAE of 0.200 and mean ice-class F1 of 76.9, confirming that the late-fusion architecture itself captures most of the gain from auxiliary environmental inputs.
Adding the SIC consistency alone reduces overall MAE to 0.197, with improvements on the thinner ice classes $C_0$--$C_2$, consistent with passive-microwave brightness temperatures being particularly informative for distinguishing multi-year from first-year ice. The environmental-compatibility regularizer alone yields a smaller MAE improvement (0.199), with its gains concentrated on $C_4$ (0.207) and $C_6$ (0.196), suggesting that ERA5 atmospheric state provides complementary context.
Combining both auxiliary regularizers achieves the best overall MAE of 0.194 and mean ice-class F1 of 77.4, achieving the lowest MAE on three of the six ice classes ($C_0$, $C_1$, $C_4$)
and matching the best result on $C_5$. The complementary class-level behavior of the two regularizers — SIC improving the thinner classes $C_0$--$C_2$ and environmental compatibility improving $C_4$ and $C_6$ — explains why their combination is better than either alone.

Table~\ref{tab:sensitivity} reports sensitivity of the Dirichlet head to its two key hyperparameters on the SAR-only configuration. The evidence scaling factor $s$ controls the magnitude of the Dirichlet concentration relative to the KL target (25\% patch budget). At $s=50$ the model under-concentrates (MAE 0.248, F1 54.5), while $s=150$ introduces over-sharpening that hurts $C_2$ and $C_3$ (MAE 0.258, F1 50.9). The default $s=100$ achieves the best MAE (0.247) and F1 (54.8), matching the target $\kappa=100$ scale as intended by the design. Replacing mean-scale aggregation with an unnormalized sum raises $C_3$ MAE from 0.320 to 0.407 and overall MAE to 0.260 with F1 dropping to 50.1, confirming that bag-size invariance is necessary for stable proportion estimates across polygons of varying patch count.
The patch-selection budget (top-\%) controls what fraction of each polygon's patches is forwarded from Module~1 to Module~2. Using all patches gives the best MAE (0.222), while a 25\% budget reduces computation but worsens MAE to 0.247 by discarding useful concentration information. A 75\% budget nearly matches the no-selection baseline (MAE 0.226, F1 61.1). Notably, the multimodal variants reach much stronger results (0.194 MAE) at only the 25\% budget, indicating that auxiliary AMSR2 and ERA5 context partly substitutes for patch quantity: with discriminative environmental signal per patch, fewer patches are needed to recover accurate polygon-level proportions. In terms of computational cost, Table~\ref{tab:efficiency} shows that reducing the patch budget from 100\% to 25\% provides approximately $3.6\times$ faster training in both configurations, $4.2\times$ ($5.7\times$) faster SAR-only (multimodal) inference, and roughly $4\times$ lower peak GPU memory.

\begin{table}[!t]
\centering
\caption{Dirichlet head sensitivity analysis (SAR-only). Per-class values are MAE.}
\label{tab:sensitivity}
\scriptsize
\setlength{\tabcolsep}{1.5pt}
\renewcommand{\arraystretch}{1.05}
\begin{tabular}{lccccccccc}
\toprule
\textbf{Configuration} & $C_0$ & $C_1$ & $C_2$ & $C_3$ & $C_4$ & $C_5$ & $C_6$ & \textbf{MAE} & \textbf{F1} \\
\midrule
\multicolumn{10}{l}{\emph{Evidence scale sensitivity ($s$) — 25\% patch budget}} \\
$s=50$                 & 0.189 & 0.216 & 0.263 & 0.350 & 0.214 & 0.254 & 0.175 & 0.248 & 54.5 \\
$s=100$ (default)      & 0.184 & 0.235 & 0.276 & 0.320 & 0.207 & 0.260 & 0.168 & \textbf{0.247} & \textbf{54.8} \\
$s=150$                & 0.178 & 0.213 & 0.313 & 0.346 & 0.206 & 0.292 & 0.158 & 0.258 & 50.9 \\
Unnormalized sum agg.  & 0.173 & 0.218 & 0.293 & 0.407 & 0.208 & 0.261 & 0.169 & 0.260 & 50.1 \\
\midrule
\multicolumn{10}{l}{\emph{Patch-selection budget sensitivity ($s=100$)}} \\
25\%                   & 0.184 & 0.235 & 0.276 & 0.320 & 0.207 & 0.260 & 0.168 & 0.247 & 54.8 \\
50\%                   & 0.186 & 0.213 & 0.270 & 0.358 & 0.205 & 0.240 & 0.147 & 0.245 & 55.4 \\
75\%                   & 0.182 & 0.225 & 0.236 & 0.325 & 0.203 & 0.184 & 0.162 & 0.226 & 61.1 \\
100\% (no clustering)  & 0.179 & 0.233 & 0.244 & 0.323 & 0.183 & 0.172 & 0.135 & \textbf{0.222} & 60.8 \\
\bottomrule
\end{tabular}
\end{table}
\begin{table}[t]
\centering
\caption{Computational efficiency at different patch-selection budgets.}
\label{tab:efficiency}
\setlength{\tabcolsep}{2.5pt}
\scriptsize
\begin{tabular}{lrrrrrr}
\toprule
& \multicolumn{3}{c}{\textbf{SAR-only}} & \multicolumn{3}{c}{\textbf{Multimodal (LF + Aux-Cons)}} \\
\cmidrule(lr){2-4}\cmidrule(lr){5-7}
\textbf{Budget} &
\textbf{Train} & \textbf{Infer.} & \textbf{GPU} &
\textbf{Train} & \textbf{Infer.} & \textbf{GPU} \\
& (s/epoch) & (ms/bag) & (GB) & (s/epoch) & (ms/bag) & (GB) \\
\midrule
25\%  & 73.1  & 44.6  & 6.5  & 88.8  & 54.0  & 7.0 \\
50\%  & 124.3 & 97.1  & 12.0 & 151.7 & 98.9  & 12.7 \\
75\%  & 165.5 & 113.7 & 16.9 & 224.3 & 142.8 & 17.9 \\
100\% & 262.6 & 188.1 & 25.3 & 317.9 & 307.2 & 25.8 \\
\bottomrule
\end{tabular}
\end{table}

Table~\ref{tab:ablation} reports component ablations on the SAR-only Dirichlet head, all evaluated using all patches. Bag-size calibration scales the Dirichlet concentration parameter by a
log-amplified size factor ($\beta=0.25$, $N_{\text{ref}}=2973$), so that larger polygons produce sharper posteriors and smaller polygons retain a stronger prior pull. This reduces MAE from 0.222 to 0.219 and raises mean ice-class F1 from 60.8 to 62.7, with the largest gains on $C_0$ and $C_1$, at the cost of small degradations on $C_2$, $C_4$, and $C_5$. Uncertainty-weighted training multiplies each bag's loss by a confidence weight derived from the Dirichlet vacuity, $\text{conf}_w = \text{clamp}(1 - K/\alpha_0,\;\text{min}=0.1)$,
down-weighting ambiguous bags early in training when the evidence head is uninformative.
This yields the best overall result, reducing MAE to 0.213 and raising mean ice-class F1 to 63.8, with notable gains on $C_3$ (0.292 vs.\ 0.323) and $C_5$ (0.166 vs.\ 0.172). Self-training with pseudo-labels blends each bag's KL target with the model's own high-confidence Dirichlet prediction after a warm-up of 30 epochs.
Bags whose total evidence $\alpha_0 \geq 150$ are assigned a blended target $\tilde{\mathbf{y}}_i = 0.5\,\mathbf{y}_i + 0.5\,\hat{\mathbf{y}}_i$, where $\hat{\mathbf{y}}_i = \boldsymbol{\alpha}_i/\alpha_{i0}$ is the model's current estimate; ambiguous bags retain their original egg-code labels.
This achieves MAE 0.216, with gains on $C_0$--$C_4$ (e.g., $C_1$: 0.211 vs.\ 0.233), but degrades the rare $C_5$ class (0.206 vs.\ 0.172) and yields lower mean F1 (61.7) than uncertainty weighting, suggesting that blending model predictions into rare-class targets can degrade class-presence detection.

\begin{table}[!t]
\centering
\caption{Dirichlet head component ablation (SAR-only), 100\% patch budget. F1 = mean F1 over ice classes (\%).}
\label{tab:ablation}
\scriptsize
\setlength{\tabcolsep}{1.5pt}
\renewcommand{\arraystretch}{1.05}
\begin{tabular}{lccccccccc}
\toprule
\textbf{Configuration} & $C_0$ & $C_1$ & $C_2$ & $C_3$ & $C_4$ & $C_5$ & $C_6$ & \textbf{MAE} & \textbf{F1} \\
\midrule
\multicolumn{10}{l}{\emph{Bag-size calibration}} \\
Without calibration              & 0.179 & 0.233 & 0.244 & 0.323 & 0.183 & 0.172 & 0.135 & 0.222 & 60.8 \\
With calibration ($\beta=0.25$)  & 0.160 & 0.217 & 0.247 & 0.317 & 0.191 & 0.182 & 0.133 & \textbf{0.219} & \textbf{62.7} \\
\midrule
\multicolumn{10}{l}{\emph{Uncertainty weighting}} \\
Without weighting                        & 0.179 & 0.233 & 0.244 & 0.323 & 0.183 & 0.172 & 0.135 & 0.222 & 60.8 \\
Uncertainty-weighted (floor$=0.1$)       & 0.175 & 0.215 & 0.245 & 0.292 & 0.186 & 0.166 & 0.134 & \textbf{0.213} & \textbf{63.8} \\
\midrule
\multicolumn{10}{l}{\emph{Self-training with pseudo-labels}} \\
Without pseudo-labels                      & 0.179 & 0.233 & 0.244 & 0.323 & 0.183 & 0.172 & 0.135 & 0.222 & 60.8 \\
Pseudo-labels ($w=0.5$, $\alpha_0\geq150$) & 0.159 & 0.211 & 0.241 & 0.305 & 0.176 & 0.206 & 0.136 & \textbf{0.216} & \textbf{61.7} \\
\bottomrule
\end{tabular}
\vspace{-3mm}
\end{table}

\section{Conclusions and Future Work}

We presented a weakly supervised framework for sea ice type proportion prediction from SAR imagery using only WMO egg-code polygon labels, without patch-level annotation.
The framework first separates water from ice, then estimates multi-label ice-type proportions within ice polygons.
The SAR-only Dirichlet model improves over the baselines by replacing fixed softmax aggregation with evidential aggregation, allowing patches to contribute according to their evidence.
Adding multimodal auxiliary context further improves performance. The late-fusion model with both auxiliary regularizers achieves the best results, with an overall MAE of 0.194 and mean ice-class F1 of 77.4\%.

Future work will extend the framework in several directions. First,
calibrating the Dirichlet evidence could provide meaningful uncertainty
estimates for identifying ambiguous polygons that require analyst review,
and could support an active-learning strategy that prioritizes uncertain
samples for expert annotation. Second, the current two-module architecture
could be extended toward joint multi-task learning with a shared feature
encoder, while graph-based spatial modeling could explicitly capture
relationships among neighboring patches. 
\begin{acks}
This research was funded by the National Science Foundation (NSF) under grant number 2531101.
\end{acks}

\bibliographystyle{ACM-Reference-Format}
\bibliography{egbib}

\appendix

\end{document}